\documentclass[11pt]{article}
\usepackage{coling2020}
\usepackage{times}
\usepackage{url}
\usepackage{latexsym}
\usepackage{graphicx}
\usepackage{amsmath}
\usepackage{xcolor}
\usepackage{float}
\usepackage{amsfonts}

\colingfinalcopy % Uncomment this line for the final submission

\title{Detecting Bias in Transfer Learning Approaches \\for Text Classification}

\author{
Irene Li\\
Department of Computer Science\\
Yale University\\
\texttt{irene.li@yale.edu} 
}

\begin{document}
\maketitle
\begin{abstract}
  Classification is an essential and fundamental task in machine learning, playing a cardinal role in the field of natural language processing (NLP) and computer vision (CV). In a supervised learning setting, labels are always needed for the classification task. Especially for deep neural models, a large amount of high-quality labeled data are required for training. However, when a new domain comes out, it is usually hard or expensive to acquire the labels. Transfer learning could be an option to transfer the knowledge from a source domain to a target domain.  A challenge is that these two domains can be different, either on the feature distribution, or the class distribution for the nature of the samples.   In this work, we  evaluate some existing transfer learning approaches on detecting the bias of imbalanced classes including traditional and deep models. Besides, we propose an approach to bridge the gap of the domain class imbalance issue. 
\end{abstract}

\section{Introduction}

Machine learning has been widely applied to multiple applications in a variety of industries. Supervised machine learning is an essential type of algorithm where both data and the corresponding labels are required when doing the process of learning.  Classification, as a very fundamental supervised machine learning task, plays an important role in many applications like face recognition and sentiment analysis \cite{le2015twitter}. However, a large amount of labeled data is often required when working with the supervised classification task, especially for deep neural networks.  When it is difficult or expensive to get enough labeled data, transfer learning \cite{sinno2010survey} is considered to be an approach to solve this issue which tries to transfer any existing knowledge to a new domain. 

Existing research has been targeting on how to improve model performance when applying transfer learning techniques from a source domain to a target domain. In the real world, there are some differences in the source and target domain. We focus on the domain class imbalance (DCI) issue, where we have different ratios of the samples for classes in two domains. For instance, imagine a binary classification problem, we have 50/50 (Pos/Neg) balanced samples in the source domain, but we may have 30/70 in the target domain. In this case, if a classifier predict all the samples to be negative cases, it can still achieve an accuracy of 0.7. But in some extreme situations like in the medical domain, we also want to examine the accuracy for the positive cases. Most of the works have shown that there is an average improvement like an average accuracy and F1 score improvement on all classes. Fewer efforts were made to check the improvements for each class, especially for some rare classes. In this work, we aim to provide some analysis on the robustness of deep transfer learning models in text classification task with a domain class imbalanced setting. This work is inspired by \cite{weiss2016investigating}, where they performed tests on some traditional transfer learning algorithms under this setting with the lack of analysis for deep models. 

\subsection{Related Work}

Most related works learn feature matching and instance reweighting for the two domains. Recent work by \cite{long2014transfer} proposed a novel Transfer Joint Matching (TJM) method to model them in a unified optimization
problem which is focusing reducing the domain difference by jointly matching the features and re-weighting the instances across domains. This method substantially improved cross-domain image recognition baseline. Another way is to learn a good feature representation shared by both two domains. \cite{Sinno2011domain} proposed a new learning method to learn the shared representation called transfer component analysis (TCA).  TAC learns some transferable feature components in a Reproducing Kernel Hilbert Space (RKHS) using Maximum Mean Discrepancy (MMD). Then in the shared feature space, some standard machine learning method can be used to do the classification task.  Similarly, \cite{long2015learning} brought this framework into a deep neural network, achieving satisfying results in classification on images. 

Adversarial learning methods are proved to be more robust when working with deep neural network models. In a recent work by \cite{tzeng2017adversarial}, a model was proposed to combine discriminative modeling, untied weight sharing, and a GAN loss, call Adversarial Discriminative Domain Adaptation (ADDA). Their results achieved a state-of-the-art unsupervised adaptation results in a classic image classification task. We will provcide more details in the following section.

The mentioned related works are mostly for CV applications, such methods are also possible to be adapted into NLP applications like sentiment classification. 

\section{Task Definition and Preliminaries}

For simplicity, first we define these notations: indices $s$ and $t$ indicate Source and Target domains. Source domain samples are $X_s=\{x_{s1},x_{s2},..,x_{sn}\}$ and the labels are $Y_s=\{y_{s1},y_{s2},..,y_{sn}\}$; similarly, target domain samples are  $X_t=\{x_{t1},x_{t2},..,x_{tm}\}$ and the labels are $Y_t=\{y_{t1},y_{t2},..,y_{tm}\}$. In our unsupervised transfer learning setting, the $Y_t$ are invisible to the model in the training stage, but $Y_s$ and $Y_t$ share the same label sets $Z=\{z_1,z_2,..,z_l\}$.
    
There are two main differences between source and target domain data. Firstly, the marginal data distributions are different, thus $P(X_s)\neq P(X_t)$. That means the features would be different for these two domains. If we take a classifier trained on source domain and directly apply it into the target, we will have a performance drop. Secondly, the sample class ratios in training sets are different. We define the sample class ratio to be the distribution of the ratio for each class. To formulate, we choose $R_s=\{r_{s1},r_{s2},..,r_{sl}\}$ to be sample class ratio of source domain, and $R_t=\{r_{t1},r_{t2},..,r_{tl}\}$ for target domain. Note that this work focus on a class imbalanced setting, in our experiments, for each domain, we choose the ratios to be different for each class. Also we want a domain class imbalanced setting which means $R_s\neq R_t $.

\subsection{Word Embeddings}
Proposed by \cite{mikolov2013distributed}, word embeddings are dense word vectors as the presentations while preserving the semantics of the words. Usually, they can be pre-trained using free text without any labels. The dimension of the dense vectors are pre-defined and then trained using the skip-gram model or cbow model \cite{mikolov2013distributed}. In this work, we will pre-train the word vectors as our input to other models.

\subsection{Generative Adversarial Networks (GANs)}

GANs \cite{goodfellow2017nips} have been widely applied in various applications and tasks and achieving promising results. It is a generalized model which contains two main components: discriminator $D$ and generator $G$. The task for the generator $G$ is to generate some cases to fool the discriminator $D$ that these cases are from real data distribution $P_{data}$. Then the discriminator $D$ is to try its best to recognize if input cases are coming from $P_{data}$. The generator $G$ takes noise variables $p_z(z)$ as input to generate sample cases. Finally, we will learn the distribution of $G$ to be $P_g$, which we are expecting it to be close to $P_{data}$. In general, it can be defined as a two-player minimax game with a value function $V(G,D)$:

\begin{equation}
\min _ { G } \max _ { D } V ( D , G ) = \mathbb { E } _ { \boldsymbol { x } \sim p _ { \text { data } } ( \boldsymbol { x } ) } [ \log D ( \boldsymbol { x } ) ] + \mathbb { E } _ { \boldsymbol { z } \sim p _ { \boldsymbol { z } } ( \boldsymbol { z } ) } [ \log ( 1 - D ( G ( \boldsymbol { z } ) ) ) ]
\end{equation}

\subsection{ADDA Model} 
As mentioned before, in transfer learning tasks, the data distributions are different. As the source domain data could be obtained more easily,  we have a good understanding of its distribution $P(X_s)$. Then we apply a feature extractor ($M_s$) for the raw inputs, then the distribution becomes $P(M_s(X_s))$ which is known already once the feature extractor ($M_s$) is well-learned with the labels provided. Then the task is, can we learn a feature extractor for the target samples ($M_t$) while the labels are missing? In this section, we introduce the idea of applying GAN to solve this issue. 

\begin{figure}
  \includegraphics[width=\linewidth]{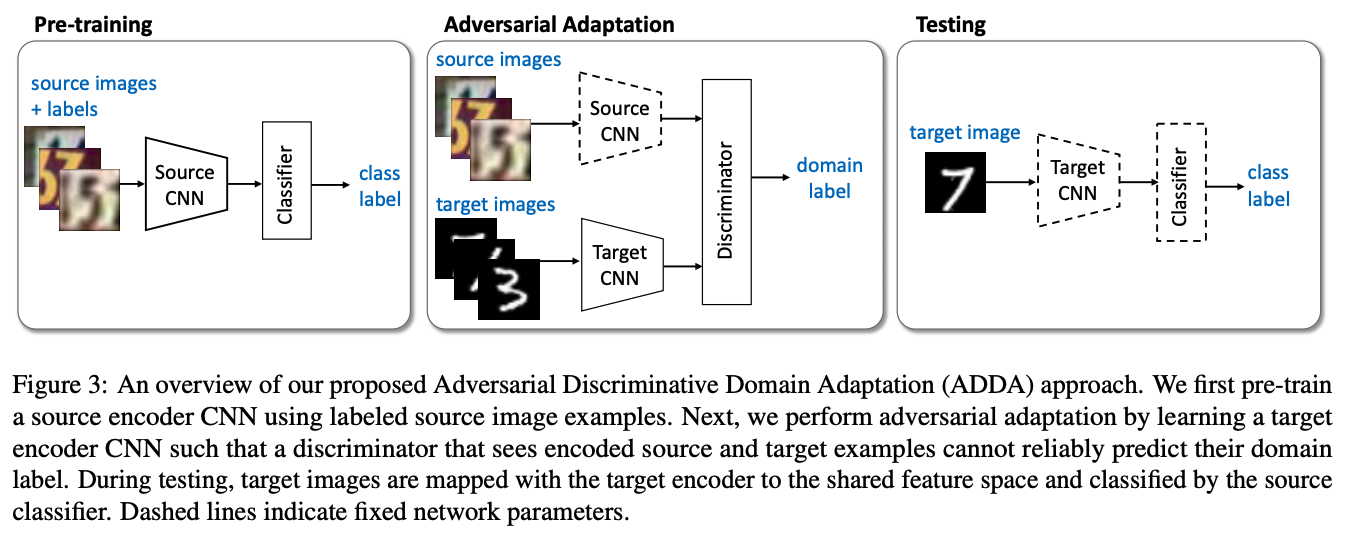}
  \caption{ADNN Model Framework, adapted from \cite{tzeng2017adversarial}.}
  \label{fig:adnn}
\end{figure}

The base model we choose is the Adversarial Discriminative Domain Adaptation (ADDA) model proposed by \cite{tzeng2017adversarial}, as shown in Figure \ref{fig:adnn}. The model was proposed to solve an unsupervised image classification task, where they have images from two domains: digit images with colorful background (source images) and hand written digit images without background (target images).

The first step is to pre-train a classifier $C$ using the source data where labels are provided. It is essential to learn the features of the training samples. The choice is to apply a Convolutional Neural Network (CNN) \cite{krizhevsky2012imagenet} model before feeding into the classifier $C$ as a feature extractor ($M_s$) for the source domain. In this step, the loss function is a typical classification loss, thus the optimization problem will be:

\begin{equation}
    \begin{aligned} \min _ { M _ { s } , C } \mathcal { L } _ { \mathrm { cls } } & \left( \mathbf { X } _ { s } , Y _ { s } \right) = \\ & - \mathbb { E } _ { \left( \mathbf { x } _ { s } , y _ { s } \right) \sim \left( \mathbf { X } _ { s } , Y _ { s } \right) } \sum _ { k = 1 } ^ { K } 1 _ { \left[ k = y _ { s } \right] } \log C \left( M _ { s } \left( \mathbf { x } _ { s } \right) \right) \end{aligned}
\end{equation}

The second step is the adversarial adaptation training process. The goal for this part is to fix the feature extractor for the source domain ($M_s$) which is learned in the fist step, and learn the one for the target domain $M_t$. The main idea is use the GAN\cite{goodfellow2017nips} framework. Then the Target CNN or the feature extractor for the target domain $M_t$ acts like a generator which tries to generate similar samples with the source images, and it takes the original target images as the input. Then a discriminator $D$ is set to determine weather the images come from source or target by outputting domain labels. In other words, we will try to approximate the distribution from target $P(M_t(X_t))$ to the known distribution from source $P(M_s(X_s))$. In this step, $M_t$ will be learned, which acts like a mapping function from the target images to source images, and then the pre-trained classifier $C$ can be adapted to the target domain images. The optimization function for the adversarial loss here can be found below. In this stage, $M_s$ is fixed while learning $M_t$. 

\begin{equation}
    \begin{aligned}
    \min _ { D } \mathcal { L } _ { \mathrm { adv } _ { D } } &\left( \mathbf { X } _ { s } , \mathbf { X } _ { t } , M _ { s } , M _ { t } \right) = \\
    & - \mathbb { E } _ { \mathbf { x } _ { s } \sim \mathbf { X } _ { s } } \left[ \log D \left( M _ { s } \left( \mathbf { x } _ { s } \right) \right) \right]
    - \mathbb { E } _ { \mathbf { x } _ { t } \sim \mathbf { X } _ { t } } \left[ \log \left( 1 - D \left( M _ { t } \left( \mathbf { x } _ { t } \right) \right) \right) \right]
    \end{aligned}
\label{eq}
\end{equation}

\begin{equation}
     { \min _ { M _ { s } , M _ { t } } \mathcal { L } _ { \mathrm { adv } _ { M } } \left( \mathbf { X } _ { s } , \mathbf { X } _ { t } , D \right) = } 
    { - \mathbb { E } _ { \mathbf { x } _ { t } \sim \mathbf { X } _ { t } } \left[ \log D \left( M _ { t } \left( \mathbf { x } _ { t } \right) \right) \right] } 
\end{equation}

The last step is the testing stage where we predict labels for the target images. We take the learned target CNN feature extractor $M_t$ from step two and the pre-trained classifier $C$ from step one to do classification. 

\section{Distance-based Backpropagation ADDA Model (DBA)}

Inspired by the work of \cite{shen2017wasserstein} and \cite{wang2019better}, it is possible to improve the model by instance re-weighting. The main idea is, we can give more weights to the instances from the target domain if they are quite similar to the instances from source domain. We can apply distance-based method to guide the process of backpropagation during training. The original stochastic gradient descent (SGD) is defined as:

\begin{equation}
    \theta \longleftarrow \theta -\alpha \nabla _{ \theta  }J(\theta ;x^{ i },y^{ i })
\end{equation}

where we have $\alpha$ as the learning rate, and the SGD step is applied during each mini-batch with $k$ instances (we set $k=10$ in all of our experiments). Then the distance-based backpropagation is defined as:

\begin{equation}
    \theta \longleftarrow \theta -\alpha \sum^{k} {w_i\nabla _{ \theta  }J(\theta ;x^{ i },y^{ i }) } 
\label{eq:weight_bp}
\end{equation}

where $w_i$ is the weight to each instance $x_i$. 

Now we show how to calculate the instance weights. The distance-based backpropagation aims to give more weights to the instances which are very close to each other from target and source domains. Here, we take the outputs of a mini batch after the feature extractors $M_s(x_s)$ and $M_t(x_t)$, and for each instance $x_{i,t}$ from target domain, the weight is defined as:

\begin{equation}
    w_i= \frac{1}{\tau Dist(M_t(x_t)-M_t(x_{i,t}) ) }
    \label{eq:weights_da}
\end{equation}

where $\tau$ is a partition function to make sure all weights (within a single mini batch) sum up to be 1. We decide the distance matric between each target domain instance output (a vector) and the output from source domain $M_s(x_s)$ (as a matrix) later. If the distance is small, then we should give more weight to that instance. It is possible to apply other distance metrics as well. 

Then the distance-based backpropagation will be applied only for the target feature extractor part. We keep the source feature extractor to be optimized via the original SGD.

\section{Experiments}

\begin{figure}
  \includegraphics[width=\linewidth]{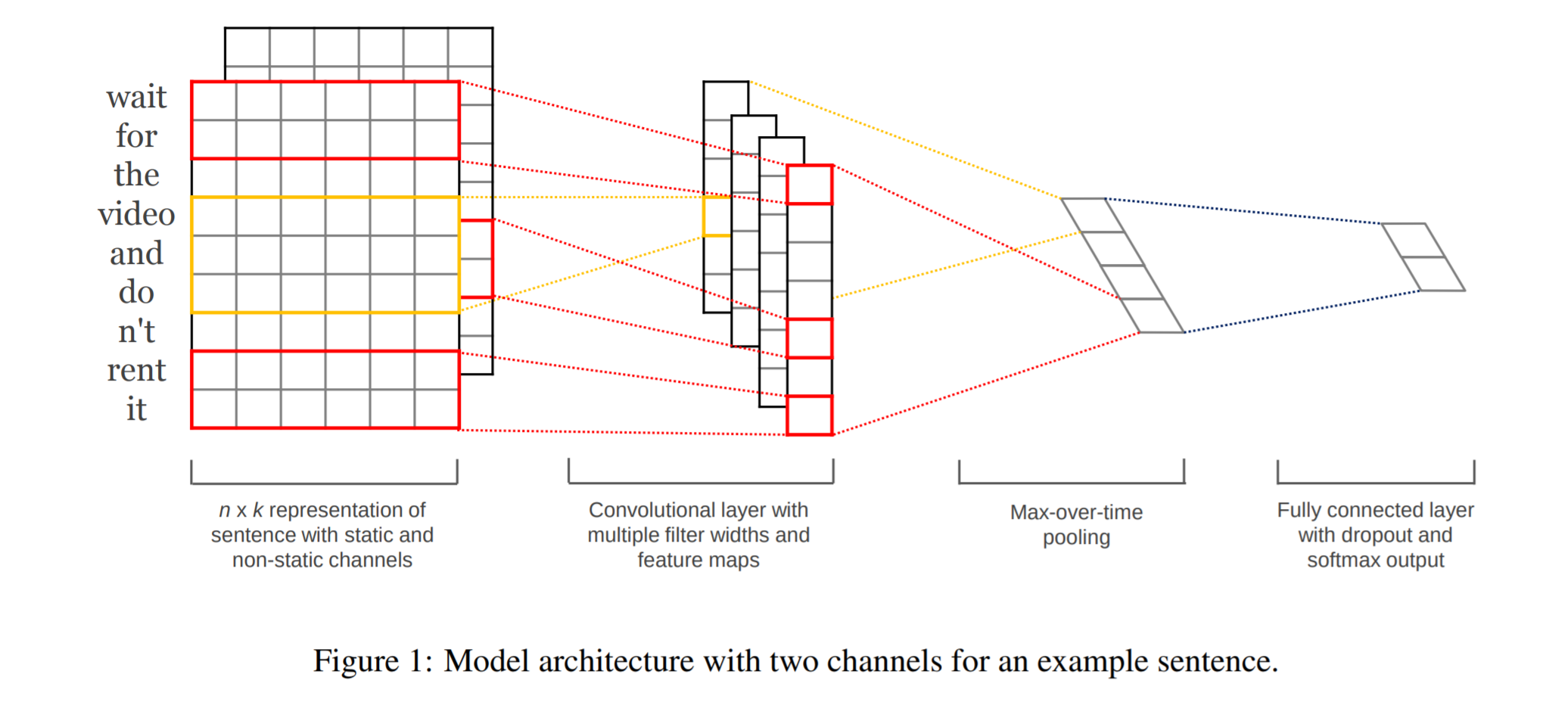}
  \caption{Convolutional Neural Networks for Sentence Classification (adapted from \cite {zhang2015sensitivity})}
  \label{fig:cnn}
\end{figure}

The original ADDA model was proposed to solve image classifications. As we are targeting on text classification task, we will choose the input to be word embeddings and CNN model to deal with the input sentences \cite{zhang2015sensitivity}. As mentioned before, we truncate each sentence to be 140 words, and each word embedding has a dimension of 128, then we make it as a 2-D matrix as an image-like input to CNN. As illustrated in figure \ref{fig:cnn}, when the word embeddings are fed into the model, then we have a convolutional layer following by a pooling layer. In the end, there is a fully-connected layer and a softmax layer to predict the labels. 

Besides, when choosing the base model for feature extractors $M_s$ and $M_t$, it is also possible to apply Long short-term memory networks (LSTM) \cite{hochreiter1997long}, but we may leave this out since CNN is already proved to be well-performed. 

Our desired algorithmic result would be a potential improvement on this existing ADDA model with an imbalanced training sample testing with text classification job. The original experimental results by \cite{tzeng2017adversarial} did not include a model performance for this particular setting. One possible idea is to apply weights for individual samples while doing the adversarial adaptation training process.  More specifically, both $\mathbb { E } _ { \mathbf { x } _ { s } \sim \mathbf { X } _ { s } } $ and $\mathbb { E } _ { \mathbf { x } _ { t } \sim \mathbf { X } _ { t } } $ in Equation \ref{eq} can be adjusted by the class ratios. For example, if the ratio is 90/10 for Pos/Neg, we may add more weights for the samples of Neg class for fairness and prevent some bias.

\subsection{Implementation}

Adapted code from \url{https://github.com/corenel/pytorch-adda}, implemented in PyTorch. Some bugs are fixed, and applied loading texts in stead of images. For simple baseline results, implemented from scratch. Use libraries include NLTK \footnote{\url{https://www.nltk.org/}} for natural language processing, and scikit-learn \footnote{\url{https://scikit-learn.org/stable/}} for machine learning methods. 

\subsection{Evaluation Metrics}

The goal is to finalize a classifier $C$ which has the ability to do classification on the target domain data $X_t$. Average accuracy and F1 scores will be reported among all the classes. Accuracy is defined as the correct number of samples classified divided by the total number of samples. F1 score is defined as:

\begin{equation}
    {\displaystyle F_{1}=\left({\frac {\mathrm {recall} ^{-1}+\mathrm {precision} ^{-1}}{2}}\right)^{-1}=2\cdot {\frac {\mathrm {precision} \cdot \mathrm {recall} }{\mathrm {precision} +\mathrm {recall} }}}.
\end{equation}

Also we check the scores for individual class. To be more specific, we define the accuracy for each class to be: the number of correct samples classified by $C$ divided by the total number of samples in that class. Similarly, we will compare F1 score for each class. By comparing the performance of different models, we are expecting an improvement in both accuracy and F1 score.

\subsection{Datasets}
We did experiments on the Multi-Domain Sentiment Dataset\footnote{\url{https://www.cs.jhu.edu/~mdredze/datasets/sentiment/}}, which contains four domains of the labeled reviews. Each domain contains 1000 positive and 1000 negative user reviews. 

\section{Results and Analysis}

\textbf{Simple Baseline methods} We tried with simple TFIDF (term frequency–inverse document frequency) features as the input to three traditional classifiers: logistic regression (\textit{LR}), random forest (\textit{RF}), naive Bayesian (\textit{NB}). Table \ref{tab:simple} shows the baseline results when we applied balanced training, please note that these are average results. Since in this method, we do not have a progress of adaptation, so column \textit{Src acc} gives the accuracy for training and testing on the source domain, while \textit{Tgt acc} gives the accuracy for training on source but testing on target data. We also provide F1 scores for both positive (marked as \textit{(p)}) and negative (marked as \textit{(n)}) class. From the results, we can see that logistic regression (\textit{LR}) performs much better than the other two classifiers, so we will then compare other model results with this method.

\begin{table}[t]
\begin{center}
\begin{tabular}{lllllll} \hline
Method & Src acc   & Tgt acc         & Src f1 (p)      & Src f1 (n)      & Tgt f1 (p)      & Tgt f1 (n)      \\ \hline
LR    & \textbf{0.8250} & \textbf{0.7294} & \textbf{0.8204} & \textbf{0.8293} & \textbf{0.7283}   & \textbf{0.7291}\\
NB   & 0.5938 &	0.5412 &	0.6011 &	0.5850 &	0.5314 &	0.5456          \\
RF    &  0.7446  &	0.6592  &	0.7569	 & 0.7304  &	0.6784  &	0.6343       \\ \hline 
\end{tabular}
\end{center}
\caption{Baseline methods}
\label{tab:simple}
\end{table}

\textbf{LR-discriminator Model} To compare with the ADDA model, it is better to keep the model structure, which is shown in Figure \ref{fig:adnn}. So we then replaced the CNN feature extractor with a traditional linear network structure and kept other layers. Specifically, the linear feature extractor is defined as:

\begin{equation}
    x'=xA^T+b
\end{equation}

where matrix $A$ and vector $b$ are trainable parameters. We use this linear layer to do feature extraction when the inputs are TFIDF features, and each token is represented by its ID. To prevent overfitting, the dimensions of each layer shrink into smaller numbers. We show the results for balanced training in Table \ref{tab:final} (\textit{LR-Dis}). The \textit{In} column shows the results for training and testing in source domain, while \textit{Out} column provides results for training in source but testing in target domain, and \textit{Adapted} is after adaptation and tested in target domain. The columns \textit{f1(p)} and \textit{f1(n)} illustrate F1 scores for positive and negative class for each corresponding method. However, this setting is not suitable for this task as most of the accuracy are lower than random guess (0.5). The bold values are the ones higher than random guess, but there is no significant improvement. A possible reason is overfitting, as the training accuracy can be 0.9.

\textbf{Deep Adaptation Adversarial Model} Results with the ADDA model are shown in Table \ref{tab:final} marked as \textit{ADDA} in the method column. Similarly, we show the accuracy in different training settings: balanced training, 1:10 (pos:neg), 3:10(pos:neg), 5:10(pos:neg) and 7:10(pos:neg). It is desired that \textit{adapted} results are higher than \textit{out-domain} results (bold values). We give in domain and out domain accuracy and f1 scores for both classes, as well as the results after adapting.

\textbf{Optimizer and Distance Metric} To select a better distance metric for calculating the weight in Equation \ref{eq:weights_da} and a good optimizer, we compare Euclidean distance (\textit{EC}) and Cosine similarity (\textit{Cosine}) in Table \ref{tab:opt}, as well as two optimizers: stochastic gradient descent (\textit{SGD}) \cite{bottou2010large} and Adam (\textit{Adam}) \cite{kingma2014adam}. It shows that by applying Adam optimizer and Cosine similarity, we could have a slightly better transferred results with a higher accuracy and f1 scores for both positive and negative class. We only show results for the experimental setting from dvd domain to the other three classes for comparison.

\begin{table}[t]
\small
\begin{center}
\begin{tabular}{lllllllllll}
\\ \hline
Source          & Target        & In & f1(p)  & f1(n)  & Out & f1(p)  & f1(n)  & Adapted & f1(p)  & f1(n)  \\ \hline
\textit{Cosine} & \textit{SGD}  &           &        &        &            &        &        &         &        &        \\
dvd             & books         & 0.725     & 0.7208 & 0.7291 & 0.688      & 0.7006 & 0.6743 & 0.5205  & 0.2696 & 0.6431 \\
dvd             & kitchen       & 0.77      & 0.7745 & 0.7653 & 0.6665     & 0.6929 & 0.6352 & 0.5123  & 0.3345 & 0.6151 \\
dvd             & electronics   & 0.78      & 0.78   & 0.78   & 0.6386     & 0.6478 & 0.629  & 0.5265  & 0.3997 & 0.6091 \\ \hline
\textit{Cosine} & \textit{Adam} &           &        &        &            &        &        &         &        &        \\
dvd             & books         & 0.7       & 0.6939 & 0.7059 & 0.7055     & 0.6892 & 0.7202 & \textbf{0.7055}  & 0.6892 & 0.7202 \\
dvd             & kitchen       & 0.755     & 0.7513 & 0.7586 & 0.6549     & 0.6702 & 0.6382 & 0.6554  & 0.6705 & 0.6389 \\
dvd             & electronics   & 0.74      & 0.7426 & 0.7374 & 0.6722     & 0.7008 & 0.6375 & \textbf{0.6722}  & 0.7008 & 0.6375 \\ \hline
\textit{EC}     & \textit{SGD}  &           &        &        &            &        &        &         &        &        \\
dvd             & books         & 0.705     & 0.6811 & 0.7256 & 0.708      & 0.6949 & 0.72   & 0.5505  & 0.5773 & 0.52   \\
dvd             & kitchen       & 0.705     & 0.6776 & 0.7281 & 0.6585     & 0.6268 & 0.6852 & 0.5696  & 0.5405 & 0.5952 \\
dvd             & electronics   & 0.775     & 0.7805 & 0.7692 & 0.6687     & 0.6927 & 0.6406 & 0.5295  & 0.4731 & 0.575  \\ \hline
\textit{EC}     & \textit{Adam} &           &        &        &            &        &        &         &        &        \\
dvd             & books         & 0.695     & 0.6115 & 0.749  & 0.65       & 0.5308 & 0.7209 & 0.65    & 0.5308 & 0.7209 \\
dvd             & kitchen       & 0.745     & 0.7437 & 0.7463 & 0.666      & 0.6549 & 0.6764 & \textbf{0.6655}  & 0.6546 & 0.6758 \\
dvd             & electronics   & 0.75      & 0.734  & 0.7642 & 0.6567     & 0.6752 & 0.6359 & 0.6567  & 0.6752 & 0.6359 \\
\hline
\end{tabular}
\end{center}
\caption{Comparison of SGD and Adam, Euclidean distance and Cosine similarity.}
\label{tab:opt}
\end{table}

\textbf{Our Model} We then apply cosine similarity and Adam optimizer as our final choice in our proposed model. Compared with other baseline models in Table \ref{tab:final}, we show that our model \textit{Our} performs better in the adapted results among all the other models.

\begin{table}[t]
\small
\begin{center}
\begin{tabular}{lllllllllll}
\\ \hline
Method   & Ratio & In & f1(p)  & f1(n)  & Out & f1(p)  & f1(n)  & Adapted & f1(p)  & f1(n)  \\ \hline
Baseline & 10:10  & 0.7211    & 0.7262 & 0.7149 & 0.6433     & 0.6460 & 0.6363 &    -    &    -   &     -  \\
LR       &        & 0.5150    & 0.4991 & 0.5261 & 0.5047     & 0.4928 & 0.5130 & 0.5033  & 0.4926 & 0.5103 \\
ADDA     &        & 0.7592    & 0.7535 & 0.7612 & 0.6660     & 0.6738 & 0.6440 & \textbf{0.6705}  & \textbf{0.6673} & \textbf{0.6723} \\
Our      &        & 0.7604    & 0.7517 & 0.7668 & 0.6703     & 0.6723 & 0.6625 & 0.6703  & 0.6724 & 0.6626\\ \hline
Baseline & 1:10   & 0.5044    & 0.6666 & 0.0337 & 0.5032     & 0.6662 & 0.0279 &    -    &    -   &     -    \\
LR       &        & 0.5067    & 0.6521 & 0.1489 & 0.5015     & 0.6495 & 0.1354 & 0.5017  & 0.6516 & 0.1236 \\
ADDA     &        & 0.5000    & 0.6667 & 0.0000 & 0.5007     & 0.6673 & 0.0000 & 0.5007  & 0.6673 & 0.0000 \\
Our      &        & 0.5000    & 0.6667 & 0.0000 & 0.5007     & 0.6673 & 0.0002 & 0.5007  & 0.6673 & \textbf{0.0002} \\\hline
Baseline & 3:10   & 0.5399    & 0.6750 & 0.1953 & 0.5208     & 0.6633 & 0.1474 &    -    &    -   &     -     \\
LR       &        & 0.5038    & 0.6151 & 0.2990 & 0.5009     & 0.6120 & 0.2989 & 0.5014  & 0.6171 & 0.2834 \\
ADDA     &        & 0.5275    & 0.6789 & 0.1015 & 0.5155     & 0.6732 & 0.0625 & 0.5104  & 0.6712 & 0.0418 \\
Our      &        & 0.5279    & 0.6794 & 0.1022 & 0.5213     & 0.6751 & 0.0834 & \textbf{0.5213}  & \textbf{0.6751} & \textbf{0.0834} \\\hline
Baseline & 5:10   & 0.6046    & 0.6905 & 0.4372 & 0.5608     & 0.6663 & 0.3303 &    -    &    -   &     -    \\
LR       &        & 0.5025    & 0.5740 & 0.3982 & 0.5055     & 0.5766 & 0.4025 & 0.5029  & 0.5791 & 0.3902 \\
ADDA     &        & 0.6338    & 0.7280 & 0.4343 & 0.5850     & 0.6980 & 0.3241 & 0.5637  & 0.6919 & 0.2481 \\
Our      &        & 0.6296    & 0.7267 & 0.4163 & 0.5787     & 0.6960 & 0.3084 & \textbf{0.5786}  & \textbf{0.6960} & \textbf{0.3083} \\\hline
Baseline & 7:10   & 0.6907    & 0.7262 & 0.6392 & 0.6077     & 0.6710 & 0.4968 &    -    &    -   &     -    \\
LR       &        & 0.5146    & 0.5592 & 0.4583 & 0.5105     & 0.5494 & 0.4624 & 0.5071  & 0.5430 & 0.4600 \\
ADDA     &        & 0.7129    & 0.7646 & 0.6267 & 0.6258     & 0.6978 & 0.4824 & 0.6363  & \textbf{0.7095} & 0.5102 \\
Our      &        & 0.7242    & 0.7610 & 0.6633 & 0.6389     & 0.6890 & 0.5369 & \textbf{0.6389}  & 0.6890 & \textbf{0.5370} \\
\hline
\end{tabular}
\end{center}
\caption{Final results.}
\label{tab:final}
\end{table}

\begin{figure}[t]
\centering
\includegraphics[width=11cm]{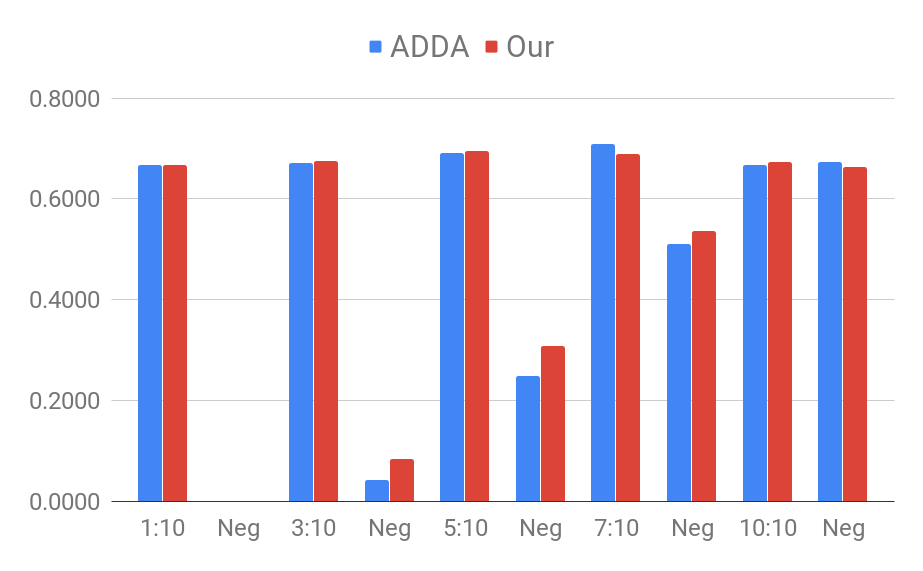}
\caption{A comparison of F1 scores among different ratio groups.}
\label{fig:final}
\end{figure}

We now compare the F1 scores of both positive and negative classes in Figure \ref{fig:final}, where x-axis shows the class with ratio group and y-axis gives the F1 scores. For example, \textit{1:10} gives the positive class under the ratio of 1:10, and the \textit{Neg} beside it indicates the negative class with the same ratio. We can see that, in a reasonable training setting, especially for 3:10, 5:10 and 7:10, our proposed model has a better F1 score in the minor class, but nearly the same F1 score with the other class. While under a fair training setting, the ADDA model performs well in both classes. While given such a small number of training and testing example, the distance metric maybe not that accurate, especially for a small batch size, which may have some impact for calculating the gradients. 

\section{Conclusion and Future Work}

We show that our proposed Distance-based Backpropagation model has the ability and potential to preventing class bias, with more unbalanced training data, the better our model perform. 

However, from our results, we show that directly applying method works in small datasets, while small datasets without pre-trianed word embeddings will easily tend to overfit. The future work is to extend our method to larger scale datasets, and it is also possible to do more analysis on other optimizers. Besides, more work can be done in solving unbalanced training issue. Our proposed approach is possible to be applied in this following manner: we assign a larger weight $w_i$ to minor class instances, and Equation \ref{eq:weights_da} then becomes:

\begin{equation}
    w_{ i }=\begin{cases} \frac { n_{ p } }{ n_{ p }+n_{ n } } \text{  if $y_i$ is negative }  \\  \\
    
    \frac {n_n}{ n_p+n_n }  \text{  if $y_i$ is positive }  \end{cases} 
\label{eq:weight_bias}
\end{equation}

In this case, we will apply Equation \ref{eq:weight_bp} to both source and target feature extractor. \\

\section*{Acknowledgements}
We thank Professor Nisheeth Vishnoi for useful discussions. 

\bibliographystyle{coling}
\bibliography{coling2020}

\end{document}